\documentclass[preprint,5p,times,twocolumn]{elsarticle}

\usepackage{amssymb}
\usepackage{amsthm}
\usepackage{amsmath}
\usepackage{multirow}
\usepackage{textcomp}
\usepackage{xcolor}
\usepackage{booktabs}
\usepackage[breaklinks,colorlinks,citecolor={blue}]{hyperref}
\hypersetup{pdfauthor={Name}}

\begin{document}

\begin{frontmatter}



\title{Fast Underwater Scene Reconstruction using Multi-View Stereo and Physical Imaging}


\author[scut]{Shuyi Hu}
\ead{ft_shuyi@mail.scut.edu.cn}

\author[scut]{Qi Liu\corref{cor1}}
\cortext[cor1]{Corresponding author.}
\ead{drliuqi@scut.edu.cn}

\affiliation[scut]{organization={School of Future Technology},
            addressline={South China University of Technology}, 
            city={Guangzhou},
            postcode={511442}, 
            country={China}}

\begin{abstract}
Underwater scene reconstruction poses a substantial challenge because of the intricate interplay between light and the medium, resulting in scattering and absorption effects that make both depth estimation and rendering more complex. While recent Neural Radiance Fields (NeRF) based methods for underwater scenes achieve high-quality results by modeling and separating the scattering medium, they still suffer from slow training and rendering speeds. To address these limitations, we propose a novel method that integrates Multi-View Stereo (MVS) with a physics-based underwater image formation model. Our approach consists of two branches: one for depth estimation using the traditional cost volume pipeline of MVS, and the other for rendering based on the physics-based image formation model. The depth branch improves scene geometry, while the medium branch determines the scattering parameters to achieve precise scene rendering. Unlike traditional MVSNet methods that rely on ground-truth depth, our method does not necessitate the use of depth truth, thus allowing for expedited training and rendering processes. By leveraging the medium subnet to estimate the medium parameters and combining this with a color MLP for rendering, we restore the true colors of underwater scenes and achieve higher-fidelity geometric representations. Experimental results show that our method enables high-quality synthesis of novel views in scattering media,  clear views restoration by removing the medium, and outperforms existing methods in rendering quality and training efficiency.
\end{abstract}




\end{frontmatter}



\section{Introduction}
\label{sec:1}

Underwater scene reconstruction is a crucial area of research with wide-ranging applications in marine science, underwater archaeology, ecology, and geoscience. Conventional methods of subaquatic investigation predominantly depend on divers and remotely operated vehicles (ROVs); however, these methodologies are frequently hindered by operational obstacles including restricted visibility, high expenses, and the requirement for specialized skills. Through the development of precise 3D depictions of submerged environments, researchers can enhance their ability to examine intricate underwater terrains, investigate marine biodiversity, and track alterations in underwater ecosystems throughout time.

Images taken underwater often experience a decrease in quality because of the distinct challenges presented by the aquatic environment. When light passes through water, it experiences attenuation that varies with distance and is sensitive to different wavelengths, as well as backscatter effects. Attenuation leads to the loss of certain colors, particularly red hue, while accentuating blue and green hues. Backscatter, in contrast, introduces a veil or haze over the image, diminishing clarity. The severity of these effects is contingent upon the proximity of the subject to the camera, in addition to its distance from the sea surface. These characteristics present a unique challenge when attempting to reconstruct the geometry of an underwater scene, due to the varying scattering properties of the medium compared to air.

In the field of Novel View Synthesis (NVS), Neural Radiance Fields (NeRF) \citep{nerf} represented scenes as continuous volumetric fields encoded by multilayer perceptions (MLP), enabling photorealistic rendering through volume rendering technique. 3D Gaussian Splatting (3D-GS) \citep{3Dgaussians} introduced differentiable rasterization for view synthesis, utilizing anisotropic 3D Gaussian primitives for explicit scene representation. Multi-View Stereo (MVS) \citep{mvsnet} methods aggregated 2D information into 3D geometric perception representations by constructing cost volumes for depth estimation. However, these aforementioned methods are typically designed for clear media such as air, where light propagation is essentially unaffected by the medium. As a result, these methods are unable to achieve satisfactory outcomes when applied directly to underwater scenes, primarily due to inaccuracies in color and density estimations while conducting 3D reconstruction.

Recently, a NeRF-based underwater scene reconstruction method, called SeaThru-NeRF \citep{seathrunerf}, was proposed, achieving state-of-the-art quality by separating the medium. However, due to the inherent limitations of the NeRF method, its training and rendering speed were too slow. An in-depth analysis of the image formation model \citep{underwatermodel} revealed that the depth prior played a crucial role. The MVSNet method, as described by \citep{mvsnet}, was a widely recognized approach for estimating depth from multiple perspectives, utilizing a cost volume generated from differentiable homologous transformations. This indicates a favorable prospect for integrating these two methodologies.

We propose a novel method that combines MVS with a physically based underwater image formation model, which is divided into two branches: one branch for depth estimation based MVS, and the other for rendering based on the physical image formation model. Depth estimation is performed using a traditional cost volume construction pipeline \citep{mvsnet}. We design a medium subnet to estimate the parameters of the imaging model and combine a color MLP for efficient rendering. By estimating the medium parameters and scene depth, we can restore the scene's true colors while improving the accuracy of its geometric representation, enabling higher-fidelity and geometrically consistent rendering from novel viewpoints. Furthermore, our proposed approach does not rely on ground-truth depth required by MVSNet \citep{mvsnet} pipeline and substantially improves both training and rendering speed when compared to SeaThru-NeRF.

In summary, the contributions of this work are shown as follows:
\begin{itemize}
\item We, for the first time, propose a novel pipeline with multi-view stereo for underwater scene reconstruction, which can synthesize novel views in scattering media and restore clear views with the medium removed.
\item We introduce a physics-based image formation model into MVS to infer the complete appearance of the scene from images without ground-truth depth, thereby enhancing the reconstruction quality of underwater environments and training efficiency.
\item We propose a network that can independently retrieve the medium parameters of the imaging model for underwater scene reconstruction.
\end{itemize}

\section{Related Work}
\label{sec:2}

\subsection{Novel View Synthesis}

NVS aims to generate new perspectives of a scene or object based on limited input views. Over the years, researchers have developed various methods to tackle this problem, spanning from conventional geometry-centered approaches to sophisticated deep learning techniques. Light field methods \citep{lumigraph, davis2012unstructured} captured dense arrays of light rays from multiple viewpoints, enabling realistic view synthesis but requiring extensive input data. Image-based rendering \citep{kalantari2016learning, riegler2020free} interpolated views through warping and blending, working well for moderate viewpoint changes. However, both approaches face challenges with sparse inputs, occlusions, and large viewpoint deviations, limiting their applicability in complex scenes. 

The advent of deep learning revolutionized NVS, enabling data-driven approaches to infer novel views directly from images \citep{BASAK2022108}. Neural representations \citep{wizadwongsa2021nex, shih20203d, lombardi2019neural, jiang2020sdfdiff} were widely used for novel view synthesis. NeRF \citep{nerf} and its derivatives marked a significant breakthrough in NVS. NeRF represented scenes as continuous volumetric fields encoded by multilayer perceptrons, enabling photorealistic rendering through volume rendering techniques \citep{barron2021mip, barron2022mip}. Efficiency improvements had been a key focus, with methods like Mip-NeRF \citep{barron2021mip} and Instant-NGP \citep{muller2022instant} significantly accelerating training and rendering using hierarchical sampling and optimized neural representations. PlenOctrees \citep{yu2021plenoctrees} and Plenoxels \citep{fridovich2022plenoxels} improved efficiency by pre-computing scene representations and dividing large-scale environments into manageable regions. Extensions for dynamic scenes, such as D-NeRF \citep{pumarola2021d} incorporated temporal information to handle motion, while NeRF-W \citep{martin2021nerf} and Block-NeRF \citep{tancik2022block} enhanced robustness in complex real-world environments. Few-shot and generalizable approaches, including PixelNeRF \citep{yu2021pixelnerf} and Depth-supervised NeRF \cite{deng2022depth}, reduced dependency on dense multi-view data, expanding NeRF's applicability to sparse-view scenarios. Notwithstanding these advancements, NeRF-based approaches still encounter difficulties with slow training and rendering times, thereby impeding their wider acceptance.

Recently, 3D-GS \citep{3Dgaussians} introduced differentiable rasterization as a novel approach to view synthesis, leveraging a set of anisotropic 3D Gaussian primitives for explicit scene representation. Its fast training speed and high-quality real-time rendering capabilities have made 3D-GS a prominent focus in the field of NVS. Many improvements have also been made to the 3D-GS method, such as antialiasing \citep{yan2024multi, yu2024mip}, effectively density control \citep{ye2024absgs} and Sparse views reconstruction \citep{charatan2024pixelsplat, yang2024gaussianobject}. However, the explicit representation of 3D-GS is not well-suited for rendering semitransparent media, such as underwater scenes with light scattering and absorption.

\subsection{Multi-View Stereo Method}

MVS aims to reconstruct a dense 3D representation of a scene from multiple views. Traditional methods \citep{fua1995object, galliani2015massively, sfm, schonberger2016pixelwise} relied on handcrafted features and similarity metrics, such as voxel-based and point-based approaches, and had laid the groundwork for this field. Voxel-based methods \citep{seitz1999photorealistic, kutulakos2000theory} evaluated the occupancy of discrete 3D units but were limited by high memory usage, while point-based methods, like PMVS \citep{pmvs}, focused on expanding reliable feature matches but may struggle in textureless regions.

Learning-based methods have become the mainstream due to their flexibility. With the development of deep learning, cost volume has been widely used for depth estimation in MVS methods \citep{ZHOU2023502}. MVSNet \citep{mvsnet} proposed for the first time an end-to-end pipeline that aggregated 2D information into a 3D geometric-aware representation by building a cost volume. The subsequent works follow the pipeline of multi-view 3D reconstruction with depth estimation from cost volume and improve from various aspects, such as improving memory consumption with recurrent plane sweeping \citep{rmvsnet, yan2020dense} or coarse-to-fine architectures \citep{cascadenet, yu2020fast, ucsnet}. However, the aforementioned MVS networks require ground truth depth as the geometric supervision for training. Thus, in the field of novel view synthesis, \citep{oechsle2021unisurf, mvsnerf, enerf, mvsgs, chen2025mvsplat} attempted to combine MVS methods with NeRF methods or 3D-GS methods. Our model extrapolates the complete appearance of the scene directly from images without the need for ground truth depth, drawing inspiration from the integration of these various methodologies.

\subsection{Image Processing in Scattering Media}

Scattering media, such as underwater environments, frequently experience challenges such as color shift, image warping, and contrast reduction, primarily caused by intricate lighting conditions involving light scattering and attenuation. Current underwater image enhancement (UIE) techniques can generally be divided into two main categories: non-deep learning-based approaches and deep learning-based approaches. CNN-based models \citep{li2021underwater, zhang2024robust, chang2025rectangling} have achieved end-to-end underwater image restoration, while Transformer-based architectures \citep{peng2023u} have further improved restoration outcomes. In contrast, non-deep learning methods are typically based on physical models and rely on prior assumptions. For example, \cite{drews2013transmission} employed the dark channel prior to estimate transmission maps specific to underwater conditions, whereas \cite{underwatermodel} refined the atmospheric scattering model for more accurate underwater image restoration. However, these non-deep learning methods rely on the accuracy of prior assumptions such as depth information, which has limited their further development. This question can be well solved by 3D vision. Various works \citep{sethuraman2023waternerf, ramazzina2023scatternerf, akkaynak2019sea} in underwater image processing have achieved remarkable results using the imaging model. The state-of-the-art NeRF-based underwater scene reconstruction method, SeaThru-NeRF \citep{seathrunerf}, incorporated the image formation model \citep{underwatermodel} into the NeRF rendering equations by estimating direct and backscatter components.

\section{Method}
\label{sec:3}

In this section, we detail the proposed MVS method for underwater scenes, which leverages an underwater image formation model to enhance MVS performance by better accounting for underwater optical characteristics. Section~\ref{sec:3.1} introduces the preliminary. The overall framework of the method is described in Section~\ref{sec:3.2}. In Section~\ref{sec:3.3}, we detail the use of the multi-view approach for depth estimation. The application of the underwater image formation model is explained in Section~\ref{sec:3.4}. Finally, Section~\ref{sec:3.5} describes the specific implementation details.

\subsection{Preliminary}
\label{sec:3.1}
\textbf{Underwater Image Formation Model.}
The image quality captured by underwater imaging systems is significantly hindered by the intricate nature of the underwater environment, predominantly because of light attenuation and scattering in water. The primary focus of underwater imaging theory is on these phenomena to elucidate and enhance the deterioration of underwater images.

The decrease in the intensity of light as it passes through water adheres to an exponential relationship, resulting from two distinct physical phenomena: absorption and scattering. The process of absorption leads to the dissipation of light energy, whereas scattering causes a redirection in the path of light propagation. Light attenuation is a multifaceted phenomenon influenced by wavelength, exhibiting varying levels of attenuation across different wavelengths, demonstrating selectivity. Red, yellow, and light green are more greatly attenuated in the visible spectrum than blue and green light, which are attenuated to a lesser degree. Consequently, underwater images generally display a blue-green coloration. The reduction of light limits the operational range of underwater imaging systems. The suspended particles and impurities found in the water are the main contributors to light scattering. Scattering effects can be classified into forward scattering and backscatter. Forward scattering pertains to the scattering of light at small angles as it reflects off the target surface before reaching the camera, causing the image to appear blurred. On the contrary, backscatter refers to light that enters the camera directly from natural or artificial light sources, following its scattering by suspended particles, causing a decrease in image contrast.

We adopt the revised underwater image formation model \citep{underwatermodel} as the general model under ambient illumination. The final image $I$ is decomposed into the direct component and the backscatter component as follows:
\begin{equation}
    I=\overbrace{ J \cdot exp({-\beta^{D}\left(\mathbf{v}_{D}\right) \cdot z})}^{\text{Direct}}+\underbrace{B^{\infty} \cdot \left(1-exp({-\beta^{B}\left(\mathbf{v}_{B}\right) \cdot z})\right)}_{\text{Backscatter}}
\end{equation}
where $J$ is the clear scene captured at depth $z$ without medium, and $B^{\infty}$ is the backscatter water color at infinity distance. The $\beta^{D}$ and $\beta^{B}$ are attenuation coefficients of the direct and backscatter components, respectively. The colors will be multiplied with attenuation coefficients to represent the effects of the medium on the color. The vectors $v_D$ and $v_B$ represent the dependencies of $\beta^D$ and $\beta^B$ on range, object reflectance, spectrum of ambient light, spectral response of the camera, and the physical scattering and beam attenuation coefficients of the water body.

\subsection{Overview}
\label{sec:3.2}

\begin{figure*}[!htb]
    \centering
    \includegraphics[width=\linewidth]{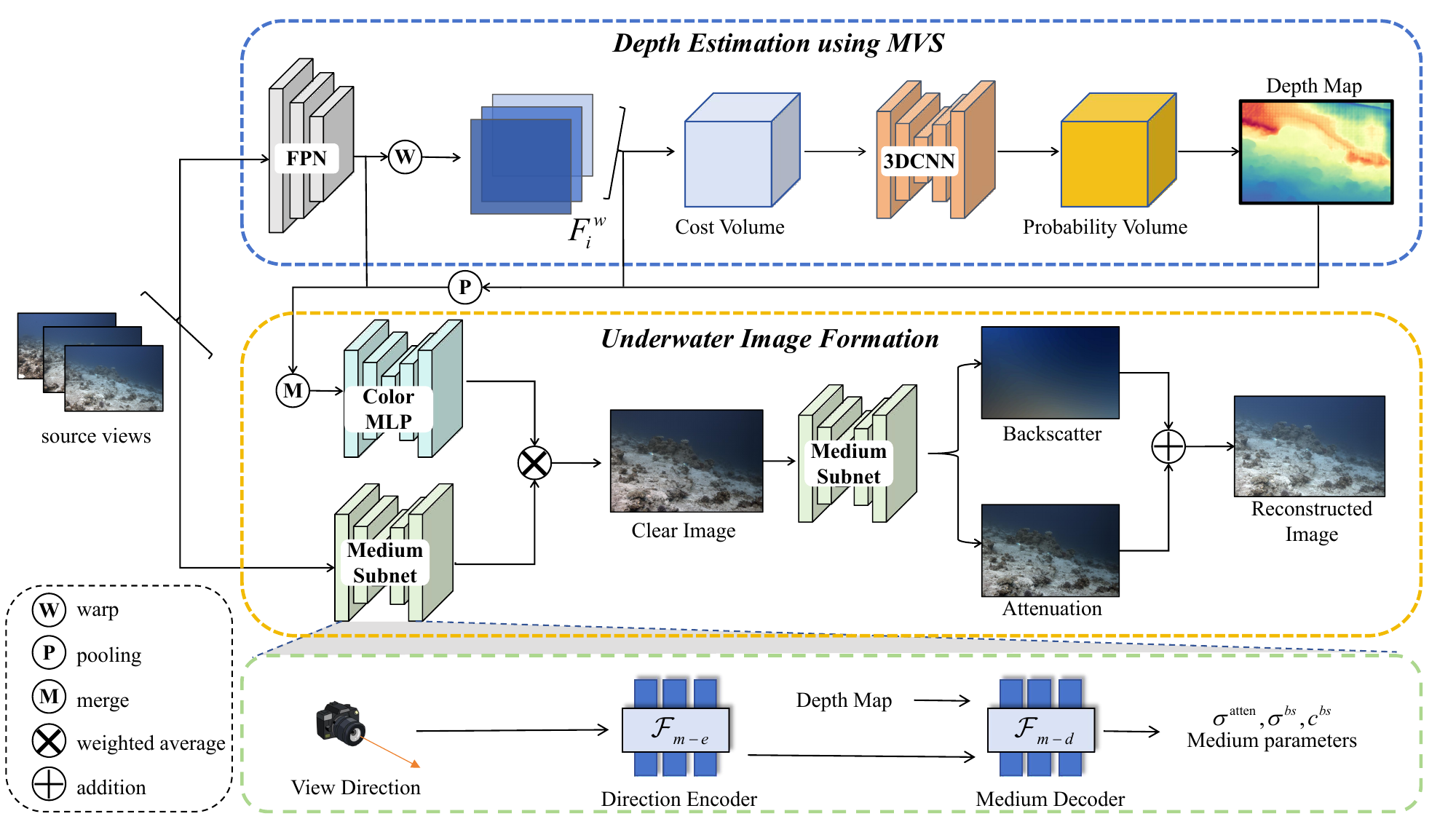}
    \caption{\textbf{Overview.} We first use an FPN to extract image features from the source view and warp them into warped features $\{F_{i}^{w}\}_{i=1}^{N}$. The distorted characteristics are combined into a cost volume, which is then processed using a 3D CNN to generate depth. Subsequently, a pooling network is utilized to consolidate features for each 3D point at the predicted depth and then integrate them with the source image features. Following this, a color MLP and a medium-sized subnet are utilized to analyze these features, resulting in a refined image with the medium removed. Ultimately, the medium subnet is utilized once more to conduct additional processing on the image, resulting in the extraction of the backscatter and attenuation images. These are subsequently combined to produce the ultimate reconstructed outcome.}
    \label{fig:pipeline}
\end{figure*}

Given multi-view images, we aim to synthesize the target image from a novel camera pose in an underwater scene. The overview of our proposed method is depicted in Fig. \ref{fig:pipeline}.  The initial step involves utilizing a Feature Pyramid Network (FPN) \citep{fpn} to extract multi-scale features from the input multi-view source views. These characteristics are subsequently transferred onto the designated camera frustum through differentiable homography to create a cost volume. This volume is then utilized by a 3D CNN for regularization to generate the depth map. The depth prediction branch of this system is constructed utilizing a cascading structure that propagates the depth map sequentially and systematically. Utilizing these depth maps, we aggregate multi-view and spatial information for each 3D point at the predicted depth by encoding features. Following this, we utilize a color MLP and the medium subnet to analyze the encoded features, which leads to the creation of a clear image with the medium removed. The intermediate subnet encodes the perspective of the view, integrates the depth map, and subsequently decodes the intermediate parameters. Ultimately, the medium subnet is utilized once more to enhance the clear image through the acquisition of backscatter and attenuation images, which are subsequently combined to produce the final reconstructed outcome.

\subsection{Depth Estimation using MVS}
\label{sec:3.3}
The depth map produced by MVS is an essential component of our workflow, facilitating seamless integration with the underwater image formation model. Our approach is grounded in the insights of learning-based MVS methods \citep{mvsnet}.

\textbf{Feature Map.}
We employ a Feature Pyramid Network (FPN) to extract multi-scale image features $\{ F_i \mid i=1, \dots ,N \}$ from input source multi-view images $\{ I_i \mid i=1, \dots ,N \}$. Specifically, each input image $I_{i} \in \mathbb{R}^{H \times W \times 3}$ is processed through the network to generate a low-resolution feature map $F_{i,1} \in \mathbb{R}^{H/4 \times W/4 \times 32}$. Subsequently, these features are fused using lateral connections and upsampling, producing two additional feature maps at higher resolutions: $F_{i,2} \in \mathbb{R}^{H/2 \times W/2 \times 16}$ and $F_{i,3} \in \mathbb{R}^{H \times W \times 8}$. This multi-resolution feature representation is then refined via smoothing convolutions to enhance the quality of the features before being passed to subsequent stages of the pipeline. The multi-scale feature extractor integrates high-resolution features with information from lower resolutions through a learned upsampling mechanism. This design ensures that each stage in the multi-stage depth prediction process uses meaningful feature representations from previous stages, making it easier to extract high-frequency features effectively \citep{ucsnet}.

\textbf{Cost Volume.}
We first partition a set of sampling planes from an initial scene range depth $\{ \mathrm{L}_i \mid i=1, \dots ,N \}$, and then warp image features of source views $F_i$ onto $D$ sweeping planes to build cost volume. This step requires the use of the differentiable homography, as described below:
\begin{equation}
    \mathrm{H}_{i}(z)=\mathrm{K}_{i}  \mathrm{R}_{i}\left(\mathrm{I}+\frac{\left(\mathrm{R}_{i}^{-1} \mathrm{t}_{i}-\mathrm{R}_{t}^{-1} \mathrm{t}_{t}\right) \mathrm{a}^{T} \mathrm{R}_{t}}{z}\right)  \mathrm{R}_{t}^{-1}  \mathrm{~K}_{t}^{-1}
\end{equation}
where $[K_i, R_i, t_i ]$ and $[K_t, R_t, t_t ]$ denote camera intrinsic, rotation and translation of input source view and target view, respectively. $I$ is the identity matrix and $a$ is the principal axis of the target view camera. We use the matrix $\mathrm{H}_i (z)$ warping the pixels at $(u, v)$ from the source view to the target view at sampled depth z. Finally, we can obtain the warped feature maps of the target view, which can be defined as:
\begin{equation}
    F_{i}^{w}(u, v, z)=F_{i} \cdot \left(\mathrm{H}_{i}(z) \cdot [u, v, 1]^{T}\right)
\end{equation}
To build the cost volume $C$, we compute the variance of these warped multi-view feature volumes $\{F_{i}^{w}\mid i=1,...,N \}$, which is widely used in MVS \citep{mvsnet, ucsnet, cascadenet} for geometry reconstruction. For each voxel in $C$, located at the coordinates $(u, v, z)$, we calculate its cost feature vector as:
\begin{equation}
    C=\frac{\sum_{i=1}^{N}\left(F_{i}^{w}-\overline{F_{i}^{w}}\right)^{2}}{N}
\end{equation}
where $\overline{F_{i}^{w}}$ is the average volume among all feature volumes. This cost volume is created utilizing variance to represent deviations in image appearance among various input perspectives. These discrepancies are a result of variations in scene geometry and view-dependent shading effects.

\textbf{Depth Prediction from Probability Volumes.}
During this step, a 3D CNN is utilized to analyze the cost volume, producing a depth probability volume $\hat{C}$. Afterwards, a depth distribution can be calculated based on this probability volume. The final depth prediction is computed by weighting each depth hypothesis using the depth probability distribution. Specifically, we obtain its probability $P_{i}$ at a specific depth plane $\mathrm{L}_{i}$ by performing the softmax operation on the depth probability volume, namely:
\begin{equation}
    P_{i} = Softmax(\hat{C})
\end{equation}
The depth value and its confidence at pixel $(u, v)$ in the target view are defined as the weighted average $\hat{L}(u,v)$ and the standard deviation $\hat{\sigma}$ are calculated by
\begin{equation}
    \hat{L}(u,v)=\sum_{i=1}^{D} P_{i}(u,v) \cdot L_{i}(u,v)
\end{equation}
\begin{equation}
    \hat{{\sigma}}(u, v)=\sqrt{\sum_{i=1}^{D} P_{i}(u, v) 
\cdot \left(L_{i}(u, v)-\hat{L}(u, v)\right)^{2}}
\end{equation}
By conducting a calculation using the depth prediction and its variance to establish a confidence interval, we can measure the uncertainty of the prediction and identify the probable range of depth within which the object is positioned. That is:
\begin{equation}
   \hat{U}(u, v)=[\hat{L}(u, v)-\lambda \hat{\sigma}(u, v), \hat{L}(u, v)+\lambda \hat{\sigma}(u, v)] 
\end{equation}
where $\lambda$ is a scalar parameter that sets the size of the confidence interval. This depth range will become the initial depth range for fine processing.

\textbf{Coarse-to-fine Prediction.}
The depth prediction pipeline is developed in a cascade structure, which allows for the propagation of the depth map in a coarse-to-fine manner. The objective is to accurately represent the geometric characteristics of the scene using multiple images and generate a detailed depth prediction map for a specific viewpoint. Subsequently, we construct a high-resolution cost volume utilizing the depth map that was previously estimated. Through the processing of this refined volume, we are able to produce a more detailed depth map and a 3D feature volume. All depth maps are stored in the buffer for utilization in the subsequent phase.

\subsection{Underwater Image Formation.}
\label{sec:3.4}

Traditional MVS methods typically use cost volume only for geometric reconstruction, but recent works \citep{mvsnerf, enerf, mvsgs} show that it can also be leveraged to infer the complete appearance of the scene. Inspired by ENeRF \citep{enerf}, we propose to apply an underwater imaging model to MVS and reconstruct the entire underwater scene.

In Section \ref{sec:3.3}, we obtain the image features $\{ F_i \mid i = 1, \dots, N \}$ from the input source views, as well as the depth maps $\hat{L}_{\text{tar}}$ for the target view and $\{\hat{L}_{\text{src}, i} \mid i = 1, \dots, N \}$ for the source views. In this section, we further process these to obtain the final reconstruction.

To effectively extract corresponding pixel-aligned features $\{ f_i \mid i=1, \dots ,N \}$ of target view from multi-view source image features, we first compute the 3D coordinates based on the depth map $\hat{L}_{\text{tar}}$ of the target view and transform them from the world coordinate system to the image coordinate systems of the source cameras. By utilizing the converted 2D coordinates, we extract features at the respective locations from the feature maps of the source images using bilinear interpolation. For every 3D point, we compute the unit direction vectors in relation to both the target camera and the source cameras.This process involves standardizing the vectors and calculating differences in ray directions to account for variations in light directions. The extracted characteristics of the sampled image are combined with the calculated ray difference vectors to create a cohesive feature representation. Finally, we used the pooling algorithm to aggregate all the pixel-aligned features to obtain the image features $f_{\text{img}}$ of the target view, namely:
\begin{equation}
    f_{\text{img}}=Pooling(f_1,\dots,f_N)
\end{equation}

Similarly, we translate the coordinates of each 3D point in the target view into a two-dimensional sampling grid, and then sample the warped multi-view feature volumes $\{F_{i}^{w} \mid i = 1, \dots, N \}$ to obtain the grid features $f_{\text{grid}}$ at the corresponding coordinates. Through the acquisition of feature representations at distinct locations, we augment the model's understanding of scene geometry. The color of a 3D point  is determined by extending the color from the originating viewpoint along the path of the destination viewpoint. Specifically, the color from the source view is modulated by a blend weight $w_{i}$, which is computed using a color MLP ${\varphi}_{color}$ via:
\begin{equation}
    w_{i}={\varphi}_{color}(f_{img}, f_{\text{grid}}, f_i)
\end{equation}
Our model is derived from the image formation model described in Sec.~\ref{sec:3.1}, the color of each pixel in both the source views and the target view adheres to the following imaging equation:
\begin{equation}
    \hat{c}=c^{\text{clr}} \cdot \exp(-\sigma^{atten} \cdot \hat{L}) + c^{\text{bs}} \cdot (1-\exp(-\sigma^{bs} \cdot \hat{L})) 
    \label{image model}
\end{equation}
where $\hat{c}$ represents the pixel color in water medium and $c^{\text{clr}}$ is the clear image  after accounting for the removal of the water medium. There are three medium parameters in the equation, $\sigma^{atten}$ and $\sigma^{bs}$ are the attenuation and scattering coefficients, respectively, and $c^{\text{bs}}$is the scattering color of the water medium.

To forecast medium parameters, a distinct subnet referred to as the `medium module' is employed to compute these parameters according to the orientation of individual pixels within an image, aligned with the world coordinate system. This module consists of an encoder and a decoder. The medium encoder employs spherical harmonic encoding  \citep{ramamoorthi2001efficient} to project the input direction map into a higher-dimensional space, thereby enhancing its ability to capture subtle directional variations and extract characteristic information across various frequencies.
\begin{equation}
    \hat{d}=\text{SHencoding}(d)
\end{equation}
where $d$ is the input direction map and $\hat{d}$ is the result after encoding. Subsequently, we put $\hat{d}$ into medium decoder to get a base output $s_{base}$. We apply different activation functions to the base output to get the final media parameters.
\begin{equation}
s_{base}=\text{MediumMLP}(\hat{d}),
\end{equation}
\begin{equation}
    \sigma^{atten},\sigma^{bs}=\text{Softplus}(s_{base}),
\end{equation}
\begin{equation}
    c^{\text{bs}}=\text{Sigmoid}(s_{base})
\end{equation}
Based on Eq. \ref{image model}, we first recover the clear image $\{c_{src,i}^{\text{clr}} \mid 1,\dots, N \}$ of source views and then perform a weighted averaging operation to obtain the clear image $c_{tar}^{\text{clr}}$ of the target view.
\begin{equation}
    c_{tar}^{\text{clr}} = \sum_{i=1}^{N} w_{i} \cdot c_{src,i}^{\text{clr}}
\end{equation}
We continue to use Eq. \ref{image model} to reconstruct the image in the water medium. Specifically, the attenuation coefficient is applied to diminish the clarity of the image in order to obtain the attenuated image, which is subsequently combined with the predicted scattering image to produce the ultimate reconstruction outcome.
\begin{equation}
    \hat{I}=I_{atten}+I_{bs},
\end{equation}
\begin{equation}
    I_{atten}= c_{tar}^{\text{clr}} \cdot \exp(-\sigma^{atten} \cdot \hat{L}_{tar}),
\end{equation}
\begin{equation}
    I_{bs}= c^{bs} \cdot (1-\exp(-\sigma^{atten} \cdot \hat{L}_{tar}))
\end{equation}
where $\hat{I}$ represents the final reconstruction result, while $I_{atten}$ and $I_{bs}$ denote the attenuated image and the scattering image, respectively.

\subsection{Loss Function}
\label{sec:3.5}

Our loss function follows the 3D-GS method with some improvements. In the initial 3D-GS method \citep{3Dgaussians}, the loss function is generally composed of an $\mathcal{L}_{1}$ Loss and a D-SSIM Loss as:
\begin{equation}
    \mathcal{L} = ( 1 - \lambda ) \mathcal{L}_{1} + \lambda \mathcal{L}_{\text{D-SSIM}}
\end{equation}
we generally set $\lambda=0.2$. Inspired by \citep{rawnerf}, applying a loss function that significantly penalizes errors in dark areas, in accordance with the way human perception condenses dynamic range, can enhance the quality of reconstructing under-illuminated scenes. To be more specific, we adopt the reconstruction loss as follows:
\begin{equation}
    \mathcal{L}_{\text {recon }}\left(\hat{C}, C^{*}\right)=\left(\frac{\hat{C}-C^{*}}{\operatorname{sg}(\hat{C})+\epsilon}\right)^{2}
\end{equation}
where $sg(\cdot)$ stands for stop gradient and $\epsilon = 10^{-3}$, $\hat{C}$ and $C^{*}$ denote the reconstructed pixel color and the ground truth pixel color. We use the reconstruction loss instead of the $\mathcal{L}_{1}$ Loss to build the proposed final loss function as:
\begin{equation}
    \mathcal{L} = ( 1 - \lambda ) \mathcal{L}_{recon} + \lambda \mathcal{L}_{\text{D-SSIM}}
\end{equation}
where $\lambda = 0.2$ is set in all our tests.

\section{Experiments}
\label{sec:4}

\subsection{Experimental Setting}
\textbf{SeaThru-NeRF Dataset.} 
SeaThru-NeRF \citep{seathrunerf} released a real forward-facing \citep{nerf} dataset consisting of four multi-view underwater scenes captured in different sea regions: IUI3 Red Sea, Cura\c{c}ao, Japanese Gardens Red Sea, and Panama. These four scenes include 29, 20, 20, and 18 images, respectively, with 25, 17, 17, and 15 images designated for training, while the remaining 4, 3, 3, and 3 images are reserved for validation. All images were captured in RAW format using a Nikon D850 SLR camera housed in a Nauticam underwater casing with a dome port, effectively minimizing refraction effects that could interfere with the pinhole camera model. The RAW images were then downsampled to an approximate resolution of 900 $\times$ 1400. Prior to further processing, the linear input images were white-balanced with a 0.5\% channel-wise clipping to eliminate extreme noise pixels. Finally, camera poses were estimated using COLMAP \citep{sfm}.

\textbf{Implementation Details.} 
Our method is implemented using PyTorch and trained on a single RTX 3090 GPU. We use the Adam \citep{adam} optimizer and train the model for 3k iterations per scene. In practice, we utilize 16 and 8 depth planes for constructing the cost volumes at the coarse and fine levels, respectively. During training, 2, 3, or 4 source views are randomly chosen as inputs with probabilities of 0.1, 0.8, and 0.1, respectively.

For the Medium Subnet, the Direction Encoder utilizes spherical harmonic encoding at level 4. The Medium Decoder is an MLP consisting of two linear layers: the first layer contains 128 hidden units, while the second layer contains 64 hidden units, both with ReLU activation. The color MLP comprises two linear layers with 24 hidden units and ReLU activation.

\textbf{Baseline Methods and Evaluation Metrics.}  
To demonstrate improvements specific to underwater scenes, we first compare our method with other MVS-based approaches, such as ENeRF \citep{enerf} and MVSGaussian \citep{mvsgs}, as well as the widely used 3D reconstruction method, 3D-GS. Most importantly, we conduct an in-depth comparison with SeaThru-NeRF \citep{seathrunerf}, the state-of-the-art NeRF-based method for underwater scene reconstruction, to emphasize the advancements introduced by our method. For evaluation, we adopt the criteria established in prior works, including ENeRF \citep{enerf}, MVSNeRF \citep{mvsnerf}, and MVSGaussian \citep{mvsgs}. For the Real Forward-facing \cite{nerf} dataset, where the peripheral regions of images are typically not visible in the input views, we focus our evaluation on the central 80\% area of the images.We use the widely-used PSNR, SSIM \citep{ssim}, and LPIPS \citep{lpips} metrics to compare the quality of synthesized views. All models are trained on the same GPU using the same data set to ensure fairness.

\subsection{Results}

\begin{table*}
\caption{\textbf{Quantitative evaluation results on the SeaThru-NeRF dataset.} $\uparrow$ indicates that larger values are better, while $\downarrow$ signifies the opposite. Bolded values highlight the best results, and underlined values represent the second-best.}
\label{tab:results}
\centering
\renewcommand\arraystretch {1}
\resizebox{1\linewidth}{!}{
\begin{tabular}{@{}lccccccccccccc@{}}
\toprule
Scene & \multicolumn{3}{c}{Cuea\c{c}ao} & \multicolumn{3}{c}{IUI3 Red Sea} & \multicolumn{3}{c}{Panama} & \multicolumn{3}{c}{Japanese Gardens Red Sea} & \multirow{2}{*}{Avg.} \\ \cmidrule(lr){2-4}\cmidrule(lr){5-7}\cmidrule(lr){8-10}\cmidrule(lr){11-13}
Methods & PSNR $\uparrow$ & SSIM $\uparrow$ & LPIPS $\downarrow$ & PSNR $\uparrow$ & SSIM $\uparrow$ & LPIPS $\downarrow$ & PSNR $\uparrow$ & SSIM $\uparrow$ & LPIPS $\downarrow$ & PSNR $\uparrow$ & SSIM $\uparrow$ & LPIPS $\downarrow$ & Time\\
\midrule
SeaThru-NeRF & \underline{32.01} & \textbf{0.964} & 0.198 & \underline{28.22} & \underline{0.937} & 0.271 & \underline{29.59} & \textbf{0.940} & 0.219 & 24.09 & 0.879 & 0.264 & 10h\\
3D-GS & 28.93 & 0.907 & \underline{0.195} & 22.59 & 0.728 & 0.351 & 29.11 & 0.894 & \textbf{0.193} & 21.50 & 0.837 & 0.2188 & 29min\\
ENeRF & 21.49 & 0.724 & 0.620 & 18.76 & 0.713 & 0.469 & 19.85 & 0.671 & 0.608 & 18.22 & 0.634 & 0.565 & 15h\\
MVSGaussian & 30.27 & 0.911 & 0.232 & 27.42 & 0.905 & \underline{0.244} & 27.38 & 0.877 & 0.237 & \underline{25.59} & \underline{0.916} & \underline{0.187} & 1h \\
Ours & \textbf{32.29} & \underline{0.942} & \textbf{0.187} & \textbf{31.82} & \textbf{0.953} & \textbf{0.241} & \textbf{29.95} & \underline{0.924} & \underline{0.206} & \textbf{26.44} & \textbf{0.940} & \textbf{0.180} & 15min\\
\midrule
\end{tabular}}
\end{table*}

\textbf{Quantitative Results.} 
To evaluate the effectiveness of our method, we compare its rendering quality against several baseline methods using the standard SeaThru-NeRF dataset. Table \ref{tab:results} presents a detailed comparison of PSNR, SSIM, LPIPS, and average training time across four distinct scenes for novel view synthesis. The results clearly demonstrate the advantages of our method, which surpasses other methods in most scenarios and shows notable enhancements in training efficiency. Our method demonstrates superior performance, resulting in an average PSNR enhancement of 1.65dB compared to the leading NeRF-based underwater scene reconstruction method, SeaThru-NeRF. In combination with the scenes in Fig. \ref{fig:results}, we can see that our improvement in PSNR is reflected mainly in two deeper scenes. Of utmost importance, our approach significantly decreases the time required for training. While our approach may not excel in all metrics, its consistent performance secured a second-place finish, closely behind the leader, yet it is noteworthy for its superior overall quality. Furthermore, our approach surpasses other MVS methods, demonstrating that the integration of MVS with a physical imaging model effectively enhances the ability to reconstruct underwater scenes.

\begin{figure}
    \centering
    \includegraphics[width=1\linewidth]{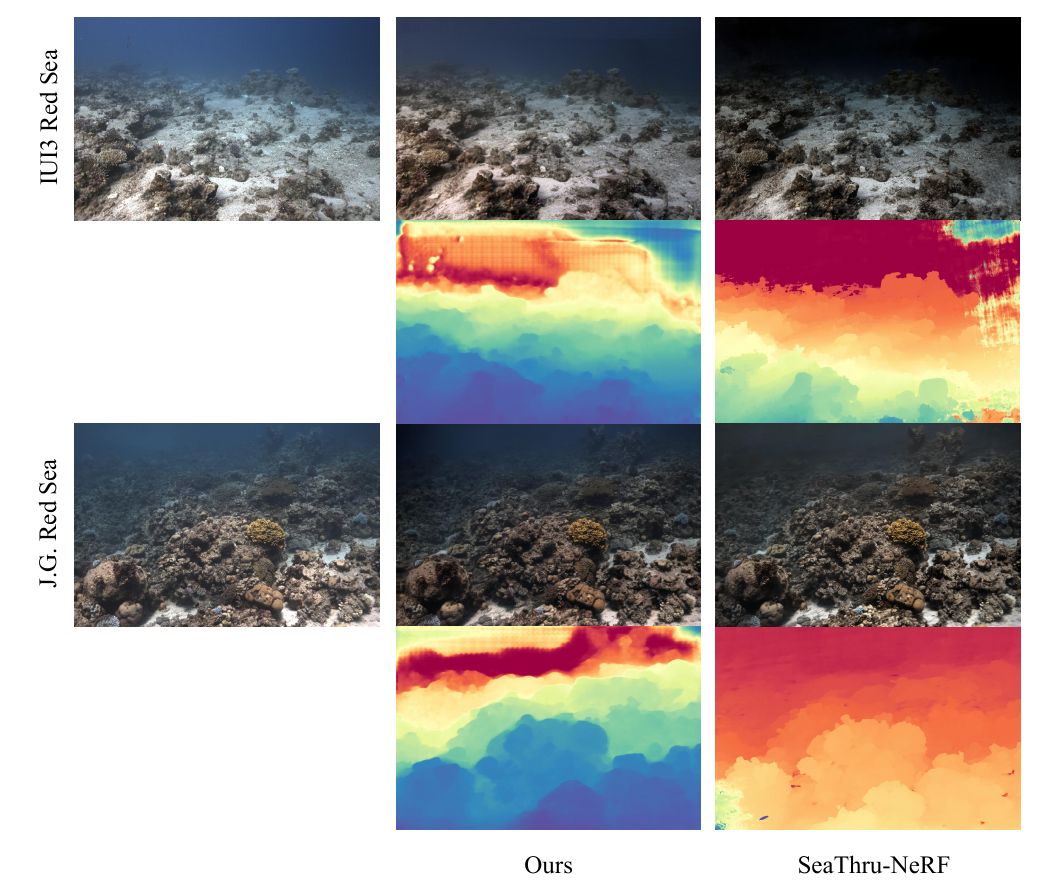}
    \caption{\textbf{Restorations and Depth Maps.} We contrast our approach with SeaThru-NeRF through the exhibition of renderings devoid of the medium. Underneath each image, we present the respective depth maps. Our restoration technique effectively preserves a greater level of color detail. In comparison to SeaThru-NeRF, the depth renderings exhibit a higher level of smoothness and coherence.}
    \label{fig:restores}
\end{figure}

\begin{figure}
    \centering
    \includegraphics[width=\linewidth]{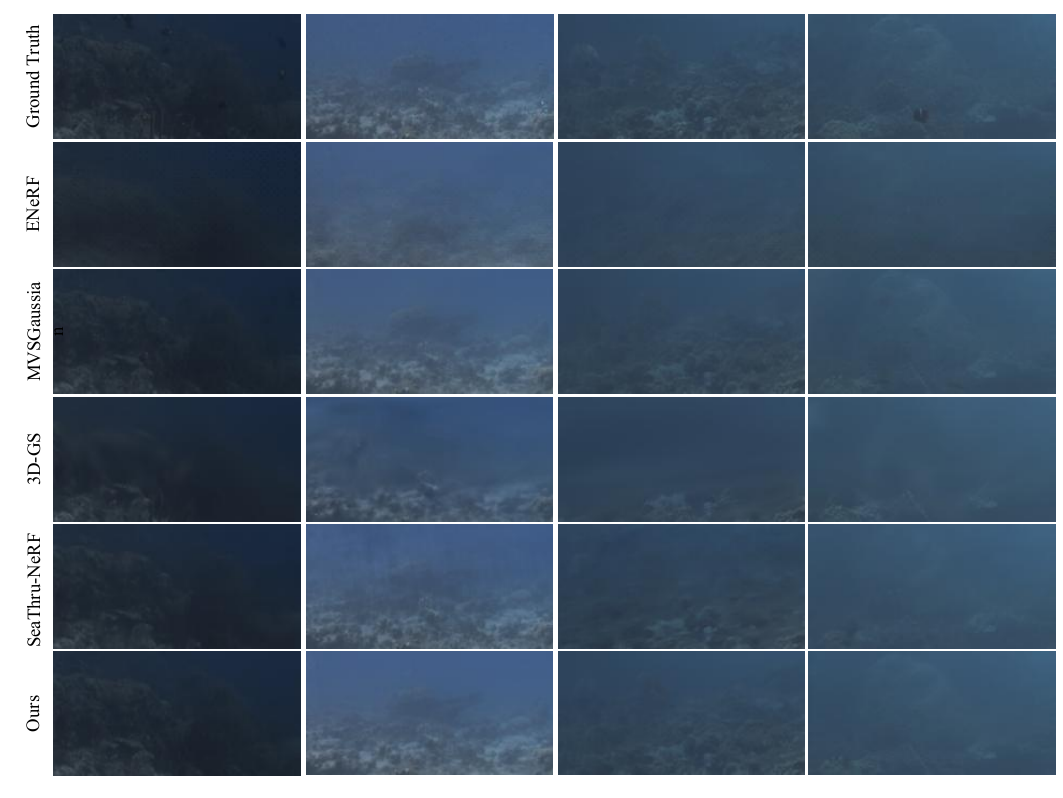}
    \caption{\textbf{The rendering of distant details.} We compare our method with several baseline methods. Our method outperforms them in rendering quality and better preserves distant geometric details.}
    \label{fig:detail}
\end{figure}

Fig. \ref{fig:results} shows the novel view renderings in the water medium. To further emphasize the superiority of our approach, we zoom in on the areas highlighted by the red squares. As shown in Fig. \ref{fig:detail}, our approach maintains sharp geometric features in remote regions, demonstrating its capacity to accurately represent intricate scene geometry, particularly in deep and complex environments. Furthermore, our approach successfully isolates objects from the scattering medium. In particular, we strive to eliminate the presence of water in the image, thereby presenting restored results that are clearer and based on the estimated medium parameters derived from our image formation model. Fig. \ref{fig:restores} demonstrates the potential of our method for restoring the color of underwater scenes. Although SeaThru-NeRF's restored images may appear grayscale, our method preserves a greater amount of color detail. Furthermore, our depth maps exhibit increased smoothness and coherence in comparison to those generated by SeaThru-NeRF, thus suggesting a higher level of accuracy in depth estimation. This is particularly advantageous for our physics-based model, as the integration of MVS's depth estimation capabilities into the physical imaging model is essential for the success of our approach.

\begin{figure*}[!htb]
    \centering
    \includegraphics[width=0.95\linewidth]{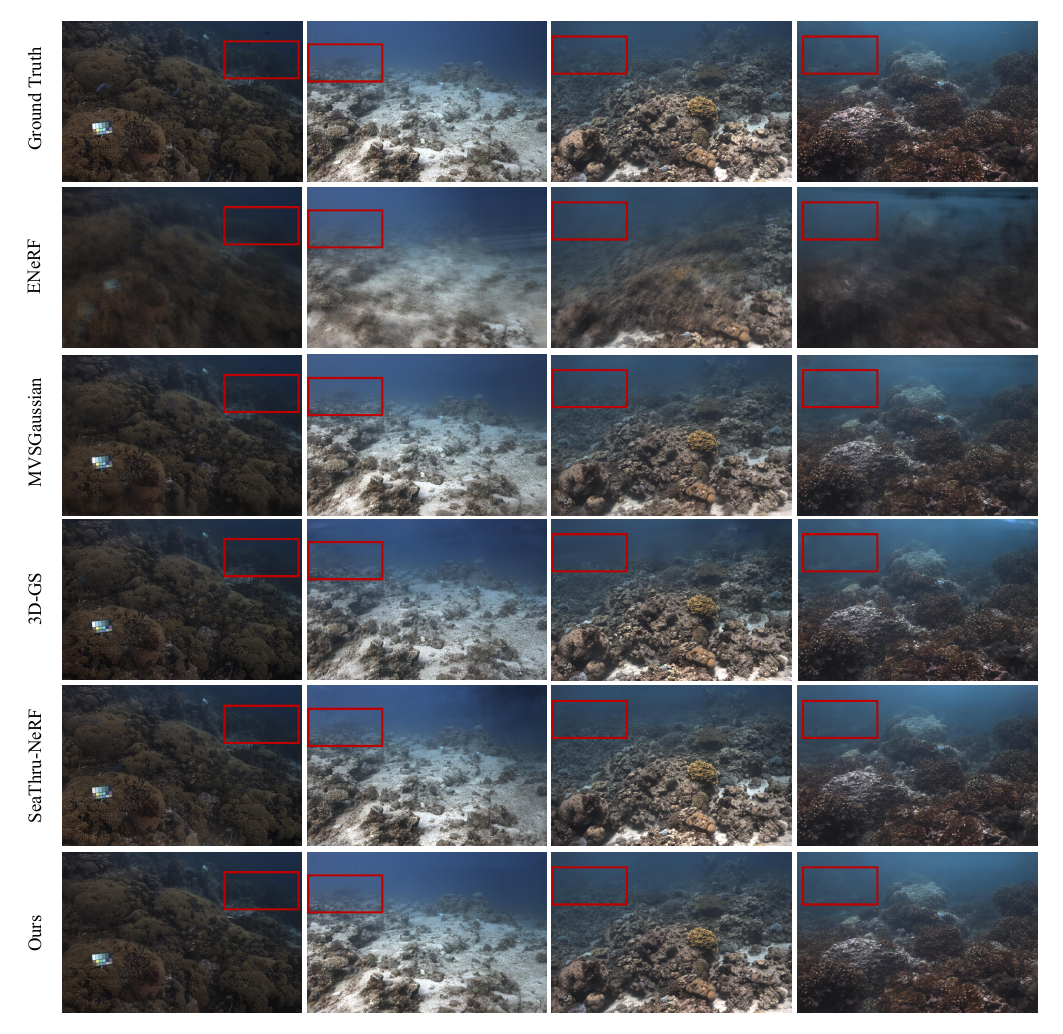}
    \caption{\textbf{Novel view synthesis in water medium.} The columns, from left to right, show the underwater scene renderings for the `Cuea\c{c}ao', `IUI3 Red Sea', `Japanese Gardens Red Sea', and `Panama' scenes. The rendering quality of distant details, highlighted within the red squares, is compared and presented in Fig. \ref{fig:detail}.}
    \label{fig:results}
\end{figure*}

\begin{table}
    \centering
    \caption{\textbf{Ablation study} on the `IUI3 Red Sea' scene.}
    \label{tab:ablation}
\begin{tabular}{ccc|c}
\hline
\textbf{Cascade} & \textbf{Recon Loss} & \textbf{Medium Subnet} & \textbf{PSNR}$\uparrow$ \\ \hline
    & \checkmark & \checkmark & 31.25\\
\checkmark &  & \checkmark & 31.73 \\
\checkmark & \checkmark  &  & 29.53 \\
\checkmark & \checkmark  & \checkmark & 31.82 \\ \hline                            
\end{tabular}
\end{table}

\textbf{Ablation study.}
We conduct ablation experiments on the `IUI3 Red Sea' scene to validate the contributions of the three key components of our framework: cascade structure, reconstruction loss, and medium subnet. As shown in Table \ref{tab:ablation}, both the cascade structure and reconstruction loss lead to improvements in reconstruction quality, as measured by PSNR. However, the greatest advantage is derived from our proposed Medium Subnet, which provides the most substantial performance enhancement. This underscores the critical role of the Medium Subnet in enhancing the effectiveness of our approach, especially when integrated with physics-based image formation models.

\section{Conclusion}
\label{sec:5}

We propose a method for underwater scene reconstruction by integrating MVS with a physical image formation model. By embedding the depth estimation capabilities of MVS directly into the imaging model, our approach achieves two key objectives: generating high-quality novel view renderings within scattering media and accurately restoring the true colors of underwater scenes.  This dual capability not only enhances the visual fidelity of rendered scenes but also provides a more accurate representation of the underwater environment, overcoming challenges associated with scattering effects and color distortion. Beyond the advancements in rendering quality, a standout feature of our methodology is the significant improvement in training efficiency than the top NeRF-based method. This efficiency enables faster processing times without compromising the accuracy or quality of the results, making our approach both practical and effective for real-world applications. In summary, the combination of superior rendering quality, efficient training, and adaptability to scattering media positions our method as a highly effective solution for underwater scene reconstruction.

Although our method represents an improvement, there are still the following directions for future improvements in underwater scene reconstruction. First, we still face challenges in handling objects with similar backgrounds in distant locations. For example, a distant blue water feature is mistakenly identified as being closer in our depth map (refer to Fig. \ref{fig:restores}). This issue is attributed in part to the feature extraction phase, especially through the FPN, which is influenced by the medium, affecting feature extraction accuracy. Future work will therefore focus on developing an improved FPN for more accurate feature extraction in underwater environments. Second, acquiring high-quality underwater scene data sets with removed water poses a challenge, leading to constraints on the quality of training data and affecting the accuracy of color recovery. In upcoming research endeavors, utilizing datasets consisting of underwater images alongside their corresponding water-removal sets is expected to significantly improve the precision of color recovery.







\bibliographystyle{elsarticle-num-names}
\bibliography{reference}

\end{document}